\documentclass[conference]{IEEEtran}
\ifCLASSINFOpdf
\else
\fi
\usepackage{graphicx}
\usepackage{fancyhdr}
\usepackage{listings} 
\usepackage{changepage}
\usepackage{cite}
\usepackage{amsmath,amsfonts}
\usepackage{algorithmic}
\usepackage{textcomp}
\usepackage{xcolor}
\usepackage{booktabs}
\usepackage[flushleft]{threeparttable}
\usepackage{makecell}
\usepackage{balance}
\usepackage{tikz}
\usetikzlibrary{shapes,arrows,decorations}
\graphicspath{ c:/user/images/ }
\usepackage{pdflscape}
\usepackage{float}
\usepackage{afterpage}
\usepackage{capt-of}
\usepackage{lipsum}
\usepackage{siunitx}
\usepackage[font=footnotesize]{subcaption}
\graphicspath{ {Figures/} }
\usepackage{placeins}
\usepackage{multirow}
\usepackage[figurename=Fig., tablename=Tab.]{caption}[2008/04/01]

\begin{document}
%
\title{ N-shot Palm Vein Verification Using Siamese Networks}

\author{\IEEEauthorblockN{Felix Marattukalam, Waleed H. Abdulla and Akshya Swain}
\IEEEauthorblockA{Department of Electrical, Computer, and Software Engineering, Auckland, New Zealand\\
felix.marattukalam@auckland.ac.nz, w.abdulla@auckland.ac.nz, a.swain@auckland.ac.nz}
}


%


\maketitle

\begin{abstract}
The use of deep learning methods to extract vascular biometric patterns from the palm surface has been of interest among researchers in recent years. In many biometric recognition tasks, there is a limit in the number of training samples. This is because of limited vein biometric databases being available for research. This restricts the application of deep learning methods to design algorithms that can effectively identify or authenticate people for vein recognition. This paper proposes an architecture using Siamese neural network structure for few shot palm vein identification. The proposed network uses images from both the palms and consists of two sub-nets that share weights to identify a person. The architecture's performance was tested on the HK PolyU multi spectral palm vein database with limited samples. The results suggest that the method is effective since it has 91.9\% precision, 91.1\% recall, 92.2\% specificity, 91.5\%, F1-Score, and 90.5\% accuracy values.
\end{abstract}


%
\IEEEpeerreviewmaketitle

\section{Introduction}

The need for contactless biometric systems have significantly increased due to the onset of the Covid-19 global pandemic. Although various \textit{extrinsic} modalities like face, iris, and palmprint \cite{fei2018feature} which are tangible part of the body are successfully being used, now there is a need for \textit{intrinsic} systems like finger vein, hand vein, and palm vein\cite{marattukalam2019palm} which are subcutaneous \cite{uhl2020handbook} and not visible to the naked eye. The features in these systems are the veins which helps liveliness detection, and is slightly more robust to spoof attacks that has been a challenge among researchers\cite{uhl2020handbook}. Palm vein systems, which is the focus of the present investigation, is an\textit{intrinsic} biometric system and preferred due to the ease of interaction with palm vein scanners. 
A palm vein biometric system has to go through several stages: \textit{acquisition}, \textit{pre-processing}, \textit{feature extraction}, \textit{decision making}. Here, we briefly look into the functions of every stage. A vein scanner captures the palm vein image using a near infrared camera in the acquisition stage. The veins are visible to the camera when illuminated under infrared light with wavelengths of 760-800 nm \cite{marattukalam2019palm}. The acquisition process suffers from issues due to uneven illumination. These issues are addressed in the pre-processing and feature extraction stages. The incoming images are cropped to a region of interest (ROI), essentially the palm region having maximum vein information to perform recognition. Then the ROI image is processed using image processing methods, and passed on to matching algorithms for decision making. The methods for matching can essentially be classified as traditional and deep learning methods. Since palm vein recognition is a classification problem and deep learning methods have been successful on such tasks, researchers are inclined towards its use.

However, the limitation that deep learning has for palm vein recognition is the need for massive databases with high quality labeled images\cite{thapar2019pvsnet}. This is often scarce in the case of palm vein recognition systems. Therefore, our work showcases the use of deep learning networks using limited samples for vein verification. The architecture used is inspired from the Siamese neural network structure, and specifically addresses the verification setting in the recognition system. 

The contributions of this research paper are: 1) The advantages of Siamese neural network architecture is exploited for palm-vein recognition by sharing the information from both the palms. As a result, a unique Siamese neural network architecture is developed for palm vein verification 2) the proposed architecture is tested on the HK PolyU multi-spectral palm vein database \cite{zhang2009online}, and its performance is evaluated. The performance evaluation show that this Siamese neural network setting is effective for palm vein verification and useful for palm vein recognition systems.

\section{Related work}

Numerous methods have been proposed to extract and match vein patterns from vein images. These patterns are used for biometric recognition using different approaches. In this section a few methods are presented for sake of completeness. 

The extraction methods can be categorised into subspace learning, local descriptor, vessel geometry, and deep learning. The recent inclination is in using Siamese networks to curb the need for large databases \cite{thapar2019pvsnet}. \cite{elnasir2014proposed} elaborates how subspace learning uses obtained coefficients as unique features for recognition. Local descriptor approaches are better described in \cite{xi2017learning}. A detailed comparison on vessel geometry is discussed in one of our previous works\cite{marattukalam2020segmentation}. Finally, deep learning approaches such as convolutional neural network(CNN), deep belief network (DBN), and auto-encoders (AE) \cite{qin2021multi} are used for feature extraction and subject recognition.

As highlighted already,deep learning in biometrics needs for large datasets . In one current public database, the images for each subject or class are limited to twelve \cite{qin2021multi} and not accounting dynamic class change. This led to the need for alternate approaches. Some researchers proposed to augment the available data. They explored  generative adversarial networks (GANs) with data augmentation to improve classification performance \cite{qin2021multi}. This did achieve reasonable performance but did not solve the problem of system speed, data privacy and storage space associated with duplication of input data. Also, data augmentation can easily lead to overfitting. Researchers are inclined towards more effective methods like similarity learning and few-shot learning. \cite{shao2021few} discusses these latest approaches and proposes a few shot learning approach for palmprint recognition, which displays good accuracy. We propose in this paper a novel combined Siamese structure for palm vein verification.

\section{Methodology}

This section discusses the database used, the general Siamese neural network architecture, and the network structure implemented with the loss functions that have been used to evaluate the network performance.  

\subsection{Database}

The palm vein image database used in our research is the HK PolyU Multispectral Palmprint and Palm Vein database (publicly available) \cite{zhang2009online} released by Hong Kong Polytechnic University (PolyU) Biometric Research Centre. A Near Infrared Region (NIR) scanner is used to capture the palm vein images. The database released has images from 250 subjects (195 male and 55 female) 20-60 years of age. The total database comprises 6000 images from 500 palms collected in two separate illumination sessions. The images captured are of resolution 352 $\times$ 288 pixels.

\subsection{Siamese Neural Networks}

\begin{figure}[ht]
\centering
    \includegraphics[width= 8cm, height=3.5cm ]{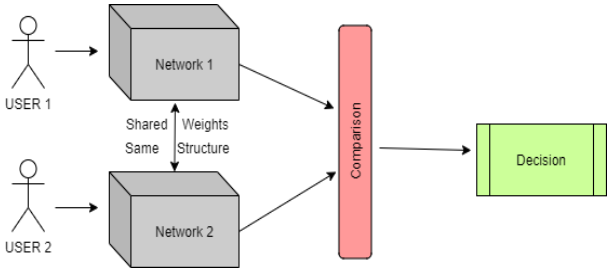}
    \caption{\label{sia} A typical Siamese neural network structure for biometric system}

\end{figure}

Even though deep learning algorithms have proven their ability to produce exceptional results, the performance of the designed algorithm is often dependant on the number of data samples available to train the network \cite{shao2021few}. 
The performance of the network improves with the increasing number of data samples. In biometric systems, especially palm vein systems, suitably labelled datasets are not readily available. One-shot or few-shot learning is the appropriate approach when only a few training examples are available for the network to train. The few-shot learning approach uses Siamese neural network. As shown in Fig. 1, a typical Siamese neural network has two paths and aims to find the similarity between its inputs. It has identical parallel networks which share the same architecture and weights. Siamese neural networks was first proposed by Bromley et al \cite{bromley1993signature} for signature verification in biometrics and is widely used in face verification tasks. Profound details about Siamese networks for image recognition are available in \cite{chicco2021siamese}.

\subsection{Network Structure}

\begin{figure*}[t]
\centering
    \includegraphics[width= 17cm, height=9cm ]{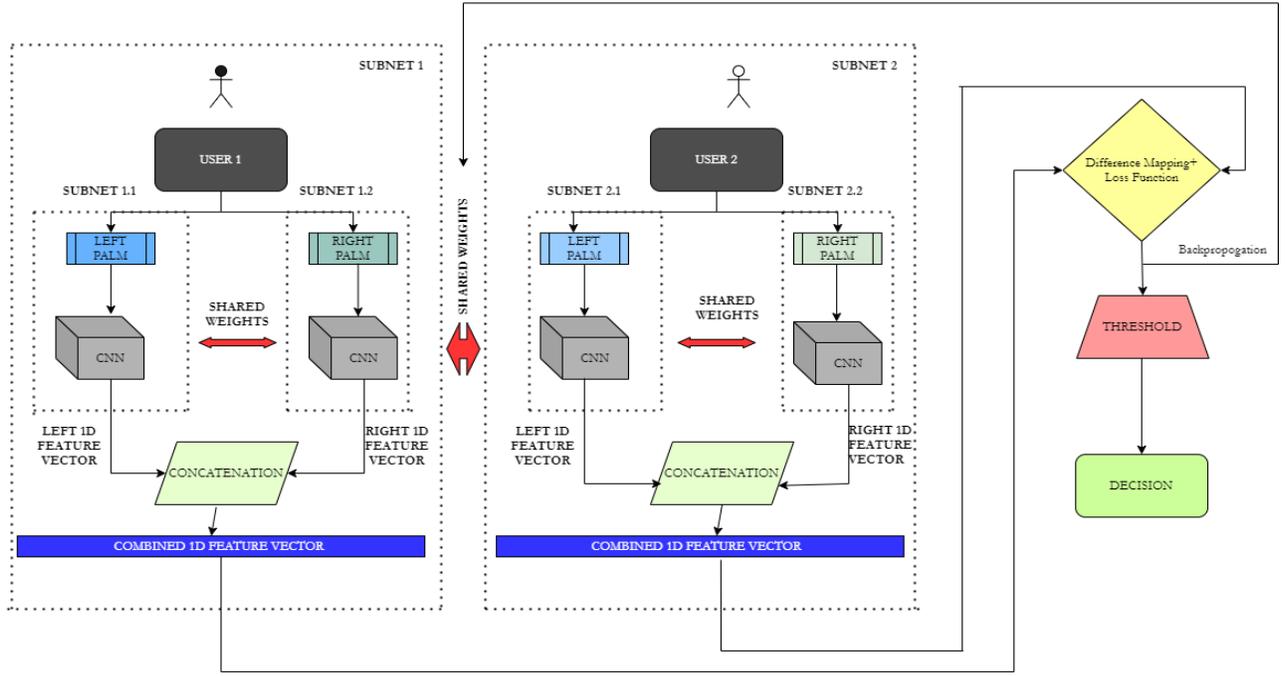}
    \caption{\label{mod}Overview of the proposed network structure based on Siamese architecture. The red arrows indicate the weights being shared between sub-networks. The sub-networks outputs are 1-D feature vectors. The distance between the feature vectors are then calculated}

\end{figure*}

This paper proposes to develop two identical networks that process two images simultaneously and compute the similarity or difference between the two images. If the images are from two different candidates, the network essentially needs to compute the similarity function and increase the distance between them. 

Fig.\ref{mod} shows the overview of the proposed network structure which is based on Siamese architecture. As introduced briefly in section 3.2, the network consists of multiple sub-networks having the same structure and share weights. Here, there are two sub-networks, subnet 1 and subnet 2. Each of them has sub subnets within them to process the input image. The left and right palm images pass through a spatial feature extractor with the convolutional neural network network structure shown in Tab. \ref{table1}. 

\begin{table}[]
\centering
\caption{CNN feature extractor structure}
\label{table1}
\begin{tabular}{|l|c|}
\hline
                                  & Input image                                 \\ \hline
\multirow{2}{*}{\textbf{Layer 1}} & Convolution 1, 64 x 3 x 3, Stride 1, Padding 0, ReLU \\ \cline{2-2} 
                                  & Batch Norm+Max Pool                      \\ \hline
\multirow{2}{*}{\textbf{Layer 2}} & Convolution 2, 64 x 3 x 3, Stride 1, Padding 0, ReLU \\ \cline{2-2} 
                                  & Batch Norm+Max Pool                      \\ \hline
\multirow{2}{*}{\textbf{Layer 3}} & Convolution 3, 64 x 3 x 3, Stride 1, Padding 1, ReLU \\ \cline{2-2} 
                                  & Batch Norm+Max Pool                      \\ \hline
\multirow{2}{*}{\textbf{Layer 4}} & Convolution 4, 64 x 3 x 3, Stride 1, Padding 1, ReLU \\ \cline{2-2} 
                                  & Batch Norm+Max Pool                      \\ \hline
\textbf{Layer 5}                  & Fully Connected, 1000 hidden units, ReLU             \\ \hline
\textbf{Layer 6}                  & Fully Connected, 128 hidden units, Sigmoid           \\ \hline
                                  & \textit{Extracted Features}                          \\ \hline
\end{tabular}
\end{table}

Consider two users, where each user submits the left and right palm image to the network. Hence, the network receives four images in total, namely, $x_1$, $x_2$ and $y_1$, $y_2$. The spatial feature extractor network  shown in Table \ref{table1} generates the feature embeddings $f(x_1)$ and $f(x_2)$ respectively, which are one-dimensional vectors of length $128 \times 1$. These vectors are then concatenated together to form $F(X)$. Similar process is followed to obtain $F(Y)$. Then the feature embeddings are subjected to a function $E$ which computes the $L_1$ distance. The function is given by eqn (\ref{eqn1}):

\begin{equation}
\centering
 E(X,Y)=d(X,Y)=||F(X)-F(Y)||
 \label{eqn1}
\end{equation}

The function $E$ will be smaller if the concatenated feature vector $F(X)$ is similar to $F(Y)$. This distance value is used to fine-tune the network weights using back propagation. A
sigmoid activation function is used to convert the distance to probability $P$.

\subsection{Loss function}

Siamese networks classify the inputs into binary classes ie. "1" being same inputs and "0" being different. Contrastive loss and binary cross-entropy function loss are the two common loss options in binary classification.

\begin{itemize} 
\item Contrastive loss:
It requires pairs of input samples. The encoder is penalized by the loss function based on the class of the input image. If the input images are from the same class, the model produces similar feature embeddings. Mathematically it is given by eqn (\ref{eqn2}):

\begin{equation}
\centering
 Loss=(1-y)*\frac{1}{2}(d)^2+(y)*\frac{1}{2} [max(0,m-d)]^2
 \label{eqn2}
\end{equation}

Here y is the actual label and will be zero when the embeddings of combined input images (left palm and right palm) and one if they are not same, d is the distance measure between the feature embeddings and the input images, m is the hyper parameter margin which is maximized if the input images are similar. If the input pairs are dissimilar and the distance d is greater than the margin $m$, no loss is incurred.

\item Binary cross-entropy loss:
It is also known as log loss and is used to calculate the classifier performance which is in the range between 0 and 1. If the predicted probability varies from actual class, the loss increases. Mathematically it is given by eqn (\ref{eqn3}):

\begin{equation}
\centering
 Loss=-y log p + (1-y) log (1-p)^2
 \label{eqn3}
\end{equation}

Here y is the class label and p is the prediction probability. It is used to differentiate between similar and different images by providing the aggregate of positive and negative loss probability.

\end{itemize}

\section{Experimental Results and Analysis}

 This section discusses about the results based on the experimental setting and the proposed architecture. The performance analysis of the results is done using the matrices namely, accuracy, precision, recall, specificity and F1-score. 

\begin{figure*}[ht]
\centering
 \includegraphics[width= 16cm, height=6cm ]{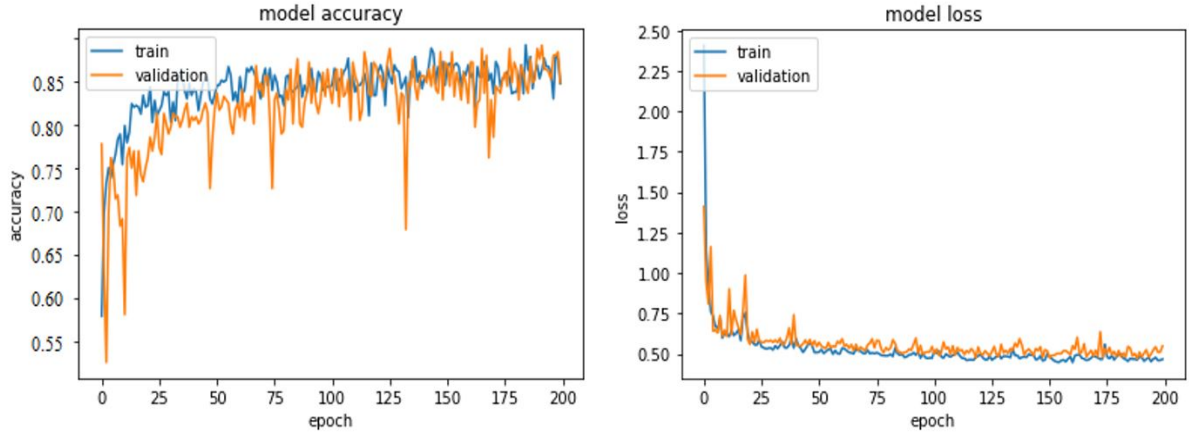}
     \caption{\label{sia} Accuracy and loss plots: Contrastive loss k=2, n=5}
 \end{figure*}

\subsection{Implementation}

This is a k-way n shot classification problem. The dataset D with a data split of 70:30 was used. The training set contains n samples from k-classes adding upto $k \times n$ samples in the training dataset and a query set in the testing dataset. Predominantly there were only two classes i.e. genuine and imposter subjects, and hence, k=2 and n varying from one to five depending on the sample set considered from the existing database. The model was trained using batch of training tasks to ultimately categorize the image during the testing task. At the end of each epoch, the model parameters were updated through back-propagation as per the loss calculated.

The HK PolyU multispectral database consists of uniform images. We used the region of interest (ROI) images of resolution 128 $\times$ 128 pixels by using the method in \cite{lin2016region}. The database was prepared into small subsets of classes containing two, three, four and five ROI images for training depending on the value of $n$. This was subjected to the spatial feature extractor described section 3.3. The experiment was carried out using the Keras framework with Tensorflow on a NVIDIA GTX 2080 8GB GPU with i7 3.3 GHz processor supported with 16 GBs of RAM. The learning rate was set at 0.0001 and Adam Optimizer was used. The one dimensional feature vector extractor which uses the structure shown in Table \ref{table1} has fully connected layers having 128 hidden units  followed by the sigmoid function.

The model was evaluated using the metrics: accuracy, precision, recall, specificity and F1- score. These parameters were preferred based on our previous study for comparison between SVM with CNN. These parameters are briefly summarised in Table \ref{table2}.

\begin{table}[h]
\caption{Performance metrics using confusion matrix}

\label{table2}
\centering
\scriptsize
\begin{tabular}{|c|c|}
\hline
\textbf{\begin{tabular}[c]{@{}c@{}} Measure\end{tabular}} & \textbf{Formula}      \\ \hline Accuracy (A)                                                           & \parbox{4cm}{\begin{equation}\frac{(TP+TN)}{(TP+TN+FP+FN)}\notag\end{equation}} \\ \hline   Precision (P)                                                          & 
\parbox{4cm}{\begin{equation}\frac{(TP)}{(TP+FP)}\notag\end{equation}}      \\ \hline       Recall (R)                                                             & \parbox{4cm}{\begin{equation}\frac{(TP)}{(TP+FN)}\notag\end{equation}}      \\ \hline       Specificity (S)                                                        & \parbox{4cm}{\begin{equation}\frac{(TN)}{(TN+FP)}\notag\end{equation}}            \\ \hline F1-Score                                                               & \parbox{4cm}{\begin{equation}\frac{(2 \times TP)}{(2 \times TP+FP+FN)}\notag\end{equation}}     \\ \hline
\end{tabular}
\end{table}
Here, the number of predictions where the classifier correctly predicts the positive class as positive is True Positive (TP), the negative class as negative is True Negative (TN), the negative class as positive is False Positive (FP), and the positive class as negative is False Negative (FN). 

\subsection{Results and Discussion}

An ideal result for this experiment would be obtaining a classification accuracy of 100 \%, precision, recall and specificity values of unity, and a 100 \% F1 score. Based on literature it can be said that such results are rare for deep learning models with small datasets. The goal of this study specifically is to exploit the benefits of Siamese neural network in palm-vein verification by discussing its performance parameters and establish its use in an end-to-end palm vein recognition system.

Experiments were performed to evaluate the network performance for different k-way,n-shot learning iterations using both contrastive and cross-entropy losses for classification. However, the results graphically represented in fig.\ref{sia} are for contrastive loss function as it was seen to be more effective than cross-entropy loss  and is in line with what has been reported in \cite{lian2018speech}. The results obtained in the experiment show that the model training and validation accuracy and loss using contrastive loss function were more stable and merged better than the plots generated for cross-entropy (even though cross-entropy used lower epochs than contrastive function). Also, Siamese neural networks use the principle of similarity between image pairs. As the experiment revolves around n-shot learning, the observations were based on how the model performance varied for different n values. As mentioned in section 4.1, the main results discussed here are for k=2 and n varying from 2 to 5. 

\begin{table}[t]
\centering
\caption{Results using contrastive loss for 2-way, n-shot settings using both palm images with n varying from 2 to 5. }
\label{table 3}
\begin{tabular}{|c|c|c|c|c|c|}
\hline
\textbf{Model} & \textbf{Accuracy} & \textbf{Recall} & \textbf{Precision} & \textbf{Specificity} & \textbf{F1-Score} \\ \hline
\textbf{k=2, n=2} & 0.862 & 0.867 & 0.874 & 0.881 & 0.871 \\ \hline
\textbf{k=2, n=3} & 0.881 & 0.885 & 0.892 & 0.899 & 0.889 \\ \hline
\textbf{k=2, n=4} & 0.892 & 0.897 & 0.906 & 0.911 & 0.903 \\ \hline
\textbf{k=2, n=5} & 0.905 & 0.911 & 0.919 & 0.922 & 0.915 \\ \hline
\end{tabular}
\end{table}

Table \ref{table 3} shows the performance metrics used. Here, contrastive loss was used and \textit{n} varies from 2 to 5. The results show that the proposed network model performance metrics increase steadily as the number of shots/ support samples is varied sequentially. This is justified because the model takes maximum benefit of the increased number of available palm vein image pairs which helps to differentiate a similar image from non-similar ones. Adam optimizer was used along with dropout prevention techniques and relevant learning rate reduction classes in Keras to pause the training process as soon as stagnancy is detected thus reducing overfitting.

\section{Conclusion}

This paper discusses the dynamics and benefits of palm vein verification and proposes a state-of-the-art deep learning Siamese neural network that can be used in contactless biometric systems. This is achieved by integrating a k-way n-shot learning network model with contrastive loss and an optimized CNN feature encoder. The results highlight that the model for k=2, n-shot learning settings using contrastive loss function is effective. The best case amongst the experiments was the 5-shot learning setting that provides an accuracy of 90.5\% in verifying the palm vein image with good recall (91.1\%) and specificity (92.2\%). These results are critical performance estimates in medical and biometric applications. The results obtained are promising owing to the fact that this model is trained with only a limited sample set of five samples from each palm for a given training class/subset.

\bibliographystyle{IEEEtran}

\bibliography{ref}

\end{document}